\documentclass[final]{cvpr}
\usepackage{cite}
\usepackage{amsmath,amssymb,amsfonts}
\usepackage{algorithmic}
\usepackage{graphicx}
\usepackage{url}
\usepackage[T1]{fontenc}
\usepackage[utf8x]{inputenc} 
\usepackage{textcomp}
\usepackage{xspace}
\usepackage{times}
\usepackage{epsfig}
\usepackage{graphicx}
\usepackage{amsmath}
\usepackage{amssymb}
\usepackage{helvet}
\usepackage{courier}
\usepackage{multirow}
\usepackage{array}

\usepackage[linewidth=1pt]{mdframed}

\def\BibTeX{{\rm B\kern-.05em{\sc i\kern-.025em b}\kern-.08em
    T\kern-.1667em\lower.7ex\hbox{E}\kern-.125emX}}

\def\eg{\emph{e.g. }} 
\def\ie{\emph{i.e. }}

\begin{document}

\global\long\def\product{\cdot}

% Paper specific

\global\long\def\centerTarget{C}

\global\long\def\ourModel{\text{our network}}

% About the classes

\global\long\def\cardinalityOfClasses{K}

\global\long\def\classIndex{k}

% Learning notions

\global\long\def\loss{L}

\global\long\def\output{y}

\global\long\def\true#1{#1_{\text{true}}}

\global\long\def\indexedTrue#1#2{#1_{\text{true},\,#2}}

\global\long\def\predicted#1{#1_{\text{pred}}}

\global\long\def\indexedPredicted#1#2{#1_{\text{pred},\,#2}}

% Hinton's notions

\global\long\def\matrixOfHinton{M}

\global\long\def\mHinton{m}

\global\long\def\row{x}

\global\long\def\dimension{n}

% Others

\global\long\def\trainingImage{X}

\global\long\def\reconstructed#1{#1_{\text{rec}}}

\global\long\def\modified#1{#1_{\text{mod}}}

\global\long\def\LTwo{L^{2}}

\global\long\def\length#1{\left\Vert #1\right\Vert }

% Electon libre / free shot

\global\long\def\FreeShot{\text{F}}

\global\long\def\numberOfFSClasses{N_{c}}

% Ponctuation within equations

\global\long\def\comma{\enspace\mbox{,}}

\global\long\def\dot{\enspace\mbox{.}}

\title{Ghost Loss to Question the Reliability of Training Data}
\author{
Adrien Deli\`ege\\
{\small University of Li\`ege}\\
\and
Anthony Cioppa\\
{\small University of Li\`ege}\\
\and
Marc Van Droogenbroeck\\
{\small University of Li\`ege}\\
}

\maketitle

\begin{mdframed}
\textbf{How to cite this work?} This is the authors' preprint version of a paper published in IEEE Access in 2020. Please cite it as follows: A. Deliège, A. Cioppa and M. Van Droogenbroeck, "Ghost Loss to Question the Reliability of Training Data", in \textit{IEEE Access}, vol. 8, pp. 44774-44782, 2020, doi: 10.1109/ACCESS.2020.2978283.
\end{mdframed}

\begin{abstract}
Supervised image classification problems rely on training data assumed
to have been correctly annotated; this assumption underpins most works
in the field of deep learning. In consequence, during its training,
a network is forced to match the label provided by the annotator and
is not given the flexibility to choose an alternative to inconsistencies
that it might be able to detect. Therefore, erroneously labeled training
images may end up ``correctly'' classified in classes which they do
not actually belong to. This may reduce the performances of the network
and thus incite to build more complex networks without even checking
the quality of the training data. In this work, we question the reliability
of the annotated datasets. For that purpose, we introduce the notion
of ghost loss, which can be seen as a regular loss that is
zeroed out for some predicted values in a deterministic way and that
allows the network to choose an alternative to the given label without
being penalized. After a proof of concept experiment, we use the ghost
loss principle to detect confusing images and erroneously labeled
images in well-known training datasets (MNIST, Fashion-MNIST, SVHN,
CIFAR10) and we provide a new tool, called sanity matrix, for
summarizing these confusions.
\end{abstract}

\section{Introduction}
\label{sec:introduction}
Large amounts of publicly available labeled training images are at
the root of some of the most impressive breakthroughs made in deep
supervised learning in recent years. From several thousands of images
(\eg MNIST~\cite{LeCun2001Gradient}, CIFAR10~\cite{Krizhevsky2009Learning},
SVHN~\cite{Netzer2011Reading}) to millions (\eg ImageNet~\cite{Deng2009ImageNet},
Quick Draw~\cite{quickdraw}), such datasets are often used as a
basis upon which new techniques are developed and compared to keep
track of the advances in the field. However, both the collection and
the annotation of such large datasets can be time-consuming and are
thus often computer-aided, if not completely automated. This results
in the production of datasets that have not been manually checked
by humans, which is a practically impossible task in many cases. Consequently,
one cannot exclude the possibility that annotation errors or irrelevant
images are present in the final datasets released publicly. These
are generally randomly split into a training set from which a network
learns to classify the images and a test set used to measure the performances
of the network on unseen images, both of which may thus contain misleading
images. By construction of these two sets, some errors can be consistent
between them, which makes them difficult to detect and may reduce
the performances when the network is deployed in a real application.
The other errors in the training and test sets contribute to decrease
the performances on the test set, which might thus appear artificially
challenging since top performances are out of reach. In all cases,
rather than questioning the reliability of the data, the usual approach
to improve the results is to design larger and deeper networks, or
networks that are able to deal with label noise~\cite{Frenay2014Classification,Patrini2017Making}.
We believe that a careful prior examination of the dataset is more
important, since it could greatly
help solving the classification task.

\begin{figure}
	\centering
	\includegraphics[width=0.95\columnwidth]{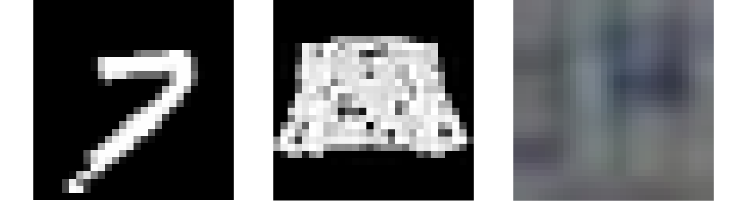}
	\caption{Training images of MNIST, Fashion-MNIST, SVHN officially labeled as
		``4'', ``shirt'', ``9''. The ghost loss helps detecting such
		problematic training images.}
	\label{fig:graphical-abstract}
\end{figure}

In this work, we introduce the principle of a ``ghost loss'' to
help analyzing four widely used training sets (MNIST, Fashion-MNIST~\cite{Xiao2017Fashion},
SVHN, CIFAR10) without requiring a manual inspection of thousands
of images. First, we detail the notion of ghost loss, which allows
the network to choose an alternative to the label provided by the
annotator without being penalized during the training phase. This
way, when an image is mislabeled, the network is given the possibility
to report that it has detected a potentially misleading image. This is done by zeroing out the loss associated with the "most likely" wrong prediction. A proof of concept experiment is presented in order to properly illustrate
its usefulness in a basic case. Then, the ghost loss is used to examine
the four previously mentioned training sets. We show that some confusions
are indeed present, such as those displayed in Figure~\ref{fig:graphical-abstract},
and we define a new tool, called sanity matrix, to summarize them.
We also show that different types of confusions can be detected. They
may originate from annotation errors, irrelevant images, or from images
that are intrinsically ambiguous and deserve multiple labels. Let us emphasize that, in this work, we focus on using the notion of ghost to analyze datasets. Further developments will be envisioned in future works.
%Let us emphasize that networks trained with a ghost loss are not meant to be evaluated on test sets, and that performance-related considerations are therefore not within our scope.
Codes will be released in due time.

\section{Related work}

Dealing with label noise in the context of supervised machine learning is a well-known issue that challenges researchers since the early developments of classifiers, as detailed in~\cite{Brodley1999Identifying}. A vast corpus of techniques has been developed, most of which are listed in a recent exhaustive survey~\cite{Karimi2020DeepLearning}, where even the latest methods related to general deep learning algorithms are indexed. Interestingly, in~\cite{Karimi2020DeepLearning}, authors propose to classify techniques for handling label noise into six (possibly overlapping) categories. In the following, we briefly describe those that are related to our work.

1. Loss functions. Many works focus on designing loss functions that are more robust to label noise than usual losses, such as~\cite{Thulasidasan2019Combating,Zhang2018Generalized}. In this category, our work is somehow close to \cite{Thulasidasan2019Combating}, which allows the network to refrain from making a prediction at the cost of receiving a penalty. A difference is that in our case, the network has to make a prediction for each class, but we zero out the loss for one of them. This principle is also close to \cite{Rusiecki2019Trimmed}, where the loss ignores the training samples with the largest loss values. However, in our case, only the prediction for one class is ignored (almost based on the loss value as well), not the training sample itself. Besides, works focusing on designing robust loss functions generally do not attempt to identify which training samples are mislabeled, contrary to our work.

2. Consistency. Some works (including ours) are based on the hypothesis that samples of the same class share similar features. Hence, the features of a mislabeled sample are not correlated with the features of the samples of its assigned class. For instance, this idea is used in \cite{Speth2019Automated} through a Siamese network trained to predict the similarity between faces, via the distance between their features in a lower-dimensional space. In comparison, our work can be seen as comparing the features with a fixed prototype of reference and thus does not need pairwise training.

3. Data re-weighting. These techniques attempt to down-weight the importance of the training samples that might be mislabeled during the learning process. The method previously mentioned \cite{Rusiecki2019Trimmed} is a particular case of this class, as some training samples are assigned a null weight. In our case, a null weight is assigned on the predictions level, hence a training sample is never completely discarded. In \cite{Wang2019Emphasis}, authors propose to re-weight the gradients computed from the predictions of the network to accomodate for the noisy labels. The weights depend on the type of loss function used and on the prediction values themselves. In our case, zeroing out the loss for one prediction also modifies the gradients but is loss-agnostic.

4. Label cleaning. Some works attempt to clean the labels on-the-fly during training. For instance, \cite{Lee2018CleanNet} use transfer learning strategies to compare the features of an image with features of reference (as for method 2.) before deciding to accept the label as is or to modify it. However, this method requires a clean subset of data, which requires careful prior examination. Our work does not require that and does not modify the labels directly. It rather helps analyzing the dataset after a network has been trained. 

Finally, let us note that no paper actually provides a way to assess the quality of a dataset as we do in this work, nor really aims at exposing mislabeled images. Besides, related works focus on reaching high performances, while we intentionnally reduce the discriminative power of our network to achieve our objectives.

\section{Method}

We assume to have a $\cardinalityOfClasses$-classes classification
network which outputs, for each input image, $\cardinalityOfClasses$
feature vectors of a given size $\dimension$. The $\classIndex\text{-th}$
feature vector is associated with the $\classIndex\text{-th}$ class
and can be seen as a vector of class-specific features used to compute
a prediction value $\indexedPredicted{\output}k$ that determines
whether the input image belongs to that $\classIndex\text{-th}$ class
or not. For instance, in~\cite{Sabour2017Dynamic}, the length of
the $\classIndex\text{-th}$ feature vector is used to compute $\indexedPredicted{\output}k$,
but other choices can be made, such as its distance from a fixed reference
vector, as in~\cite{Deliege2018HitNet}. Most standard architectures can be seen as using $n=1$ and
the $k$-th ``feature vector'' is $\indexedPredicted{\output}k$
itself. As for any network, the prediction vector $\predicted{\output}$
made of the $\cardinalityOfClasses$ values obtained from the $\cardinalityOfClasses$
feature vectors is compared with a one-hot encoded ground-truth vector
$\true{\output}$ through a loss function $\loss(\true{\output},\predicted{\output})$.
The $\classIndex\text{-th}$ component of $\true{\output}$ is noted $\indexedTrue{\output}{\classIndex}$.
We assume that the loss can be decomposed as the sum of $\cardinalityOfClasses$
losses, in a multiple binary classifiers fashion: 
\begin{multline}
\loss(\true{\output},\predicted{\output})=\sum_{\classIndex=1}^{\cardinalityOfClasses}\indexedTrue{\output}{\classIndex}\,L_{1}^{k}(\indexedPredicted{\output}{\classIndex})\\
+(1-\indexedTrue{\output}k)\,L_{0}^{k}(\indexedPredicted{\output}{\classIndex})\comma\label{eq:loss-init}
\end{multline}
where $L_{1}^{k}(\indexedPredicted{\output}{\classIndex})$ (resp.
$L_{0}^{k}(\indexedPredicted{\output}{\classIndex})$) is the loss
generated by $\indexedPredicted{\output}k$ when $\indexedTrue{\output}{\classIndex}=1$
(resp. $0$). One generally chooses the same $L_{1}^{k}(.)$ for all
the values of $\classIndex$, up to class-specific multiplicative
factors used to handle class imbalance; the same principle holds for
$\loss_{0}^{k}(.)$. For example, in the case of the categorical cross-entropy loss preceded by a softmax activation, one has $n=1$, $L_{1}^{k}(.)=-\log(.)$ for all $\classIndex$ and $L_{0}^{k}(.)=0$
for all $\classIndex$.

In this configuration, the network is not allowed to question the
true labels $\true{\output}$ provided by the annotator. Indeed, in
the back-propagation process, in order to minimize the loss, it is
forced to update all the feature vectors towards given target zones,
\ie zones where their associated prediction values are close to (or
far from) target values such as $0$, $0.5$, $1$, ...~. However,
this can be problematic in some cases. For example, if an image of
a horse is erroneously labeled as a truck, then the network has to
find a way to produce a feature vector ``horse'' whose prediction
value indicates that the image is not of a horse and a feature vector
``truck'' whose prediction value indicates that the image represents
a truck. Now, considering that the network has many other training
images of horses and trucks correctly labeled at its disposal, it
will be able to detect their characteristics and to produce discriminative
``horse'' feature vectors for horses and ``truck'' feature vectors
for trucks. Therefore, despite being labeled as truck, the erroneously
labeled image of a horse is still likely to exhibit a horse feature
vector giving a prediction value which is much ``more significant''
than the prediction values of the airplane, frog, or any other class.
By ``more significant'', we mean closer to the aforementioned target
value, \eg ``more likely'' in a probabilistic setting. In such
a case, it would be more appropriate not to penalize the horse feature
vector, which conveniently already appears to be associated with ``the
most significant non true class''. This is the motivation behind
the present work.

\paragraph*{Definitions. }

\begin{figure}
	\centering
	\includegraphics[width=0.99\columnwidth]{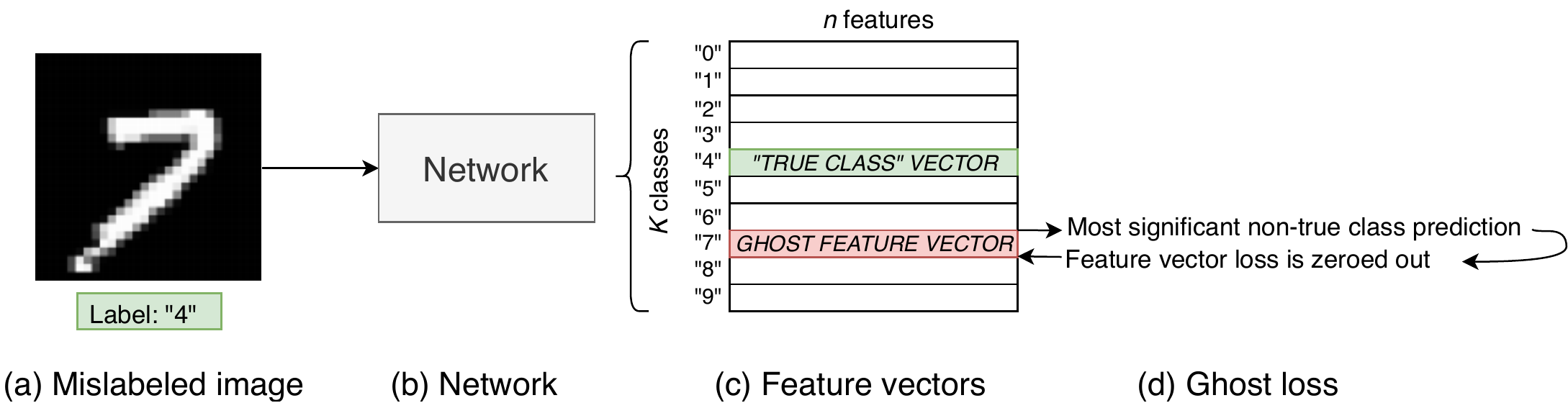}
	\caption{Illustration of the \emph{ghost feature vector} and \emph{ghost loss}
			principle. A training image (a) of a digit ``7'' is mislabeled
			as ``4''. The network (b) outputs feature vectors (c). The feature
			vector leading to the most significant prediction value among those
			different from the feature vector ``4'' is selected as the ghost
			feature vector. Here, the feature vector ``7'' is expected to be
			the ghost feature vector. The loss generated by this ghost feature
			vector is then zeroed out (d).}
	\label{fig:ghost-capsules}
\end{figure}

In order to give the network the capacity to opt for an alternative
to the labels provided by the annotator, we introduce the notions
of ``\emph{ghost feature vector''} and ``\emph{ghost loss''. }The
ghost feature vector of an image is the feature vector corresponding
to the most significant prediction value among the $\cardinalityOfClasses-1$
feature vectors not corresponding to the feature vector of the true
class. The contribution to $L$ engendered by the ghost feature vector
is zeroed out and is called a ghost loss:
\begin{equation}
L_{0}^{k_{\text{ghost}}}(\indexedPredicted{\output}{\classIndex_{\text{ghost}}})\leftarrow0\dot
\end{equation}
This means that, during the training phase, the ghost feature vector
is not forced to target any particular zone, and is not involved in
the update of the weights from one batch to the next; it is essentially
invisible in the back-propagation, hence its name. The concept of
ghost vector is illustrated in Figure~\ref{fig:ghost-capsules}.

From an implementation point of view, training a network with ghost
feature vectors and a ghost loss is similar to training it without,
only the loss defined in Equation~\ref{eq:loss-init} needs to be
adjusted. Given a one-hot encoded label $\true{\output}$ and the
prediction vector of the network $\predicted{\output}$, the loss
$L_{\text{ghost}}$ can be written as: 
\begin{multline}
\loss_{\text{ghost}}(\true{\output},\predicted{\output})=\sum_{\classIndex=1}^{\cardinalityOfClasses}\indexedTrue{\output}{\classIndex}\,L_{1}^{k}(\indexedPredicted{\output}{\classIndex})\\
+{\color{blue}\gamma}_{{\color{blue}k}}(1-\indexedTrue{\output}k)\,L_{0}^{k}(\indexedPredicted{\output}{\classIndex})\comma\label{eq:loss-ghost}
\end{multline}
where $\gamma_{k}=0$ if $\classIndex=k_{\text{ghost}}$, and $\gamma_{k}=1$
otherwise.

Two important characteristics are thus associated with a ghost feature
vector: its class $k_{\text{ghost}}$, which is always one of the
$\cardinalityOfClasses-1$ classes not corresponding to the true class
of the input image, which we call the \emph{ghost class,} and its
associated prediction value $\indexedPredicted{\output}{\classIndex_{\text{ghost}}}$,
that we call \emph{ghost prediction}. The ghost class $k_{\text{ghost}}$
of an image is obtained in a deterministic way, at each epoch, as
the class of the most significant non true class prediction; it is
not a Dropout \cite{Srivastava2014Dropout}, nor a learned Dropout
variant (as \eg \cite{Lee2017DropMax}). The ghost class of an image
can change from one epoch to the next as the network trains. Ideally,
a ghost feature vector will have a significant ghost prediction when
its ghost class is a plausible alternative to the true class or if
the image deserves an extra label, and will not otherwise. The evolution
of a ghost feature vector during the training is dictated by the overall
evolution of the network. 

Subsequently, in the situation described above, at the end of the
training, the image of a horse labeled as a truck should actually
display two significant prediction values: one for the feature vector
corresponding to the ``truck'' class since the network was forced
to do so, and one for the feature vector corresponding to the ``horse''
class. Indeed, the network probably selected the ``horse'' feature
vector as ghost feature vector because the image displays the features
needed to identify a horse and it was not forced to give it a non
significant prediction value. Looking at the images with significant
prediction values at the end of the training allows to detect the
images for which the network suspects an error in the label, or for
which assigning two labels might be relevant.

We can derive a new tool from the notions defined above that helps
aggregating information on the number of training images that might
be considered as suspicious. This tool takes the form of a matrix,
which we call \emph{``sanity matrix''}, in which the counts of the
pairs (true class, ghost class) are reported when the ghost prediction
is sufficiently significant with respect to a given threshold. 

Let us note that the ghost loss principle should not be used along
with the categorical cross-entropy loss preceded by a softmax activation.
Indeed, in such a case, the use of $\loss_{0}^{k}$ is obsolete since
the cross-entropy loss can be seen as a particular case of Equation~\ref{eq:loss-init}
in which $L_{1}^{k}(.)=-\log(.)$ for all $\classIndex$ and $L_{0}^{k}(.)=0$
for all $\classIndex$. Therefore, the introduction of $\gamma_{k}$
in Equation~\ref{eq:loss-ghost} has no impact on the loss.

\section{Experiments}

\subsection{Network used}

\begin{figure}
	\centering
	\includegraphics[width=0.99\columnwidth]{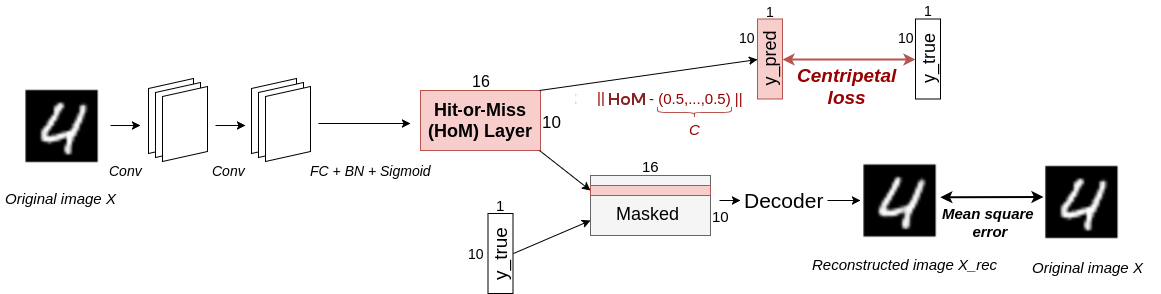}
	\caption{Graphical representation of the network structure used for the experiments
		in this paper.}
	\label{fig:Hitnet}
\end{figure}

The network used in the following experiments comes from~\cite{Deliege2018HitNet}. It is a simplified and adapted version of the
network introduced in~\cite{Sabour2017Dynamic}. %The ghost loss could be applied with CapsNet but we simplify it to speed up its training and to keep only the relevant parts for the present work.
The network that we use for the experiments is represented in Figure~\ref{fig:Hitnet}
and is composed of the following elements, as in~\cite{Deliege2018HitNet}. 

First, two $9\times9$ (with strides (1,1) then (2,2)) convolutional
layers with 256 channels and ReLU activations are applied, to obtain
feature maps. Then, we use a fully connected layer to a $\cardinalityOfClasses\times\dimension$
matrix, followed by a BatchNormalization and an element-wise sigmoid
activation, which produces what we call the Hit-or-Miss (HoM) layer
composed of $\cardinalityOfClasses$ feature vectors of size $\dimension$.
We set $\dimension=16$ as in~\cite{Deliege2018HitNet}. The Euclidean
distance with the central feature vector $\centerTarget:(0.5,\,\ldots,\,0.5)$
is computed for each feature vector of HoM, which gives the prediction
vector of the model $\predicted{\output}$. This is done such that,
for a given image, the feature vector of the true class is close to
$\centerTarget$ (prediction value $\indexedPredicted{\output}k$
close to $0$) and the feature vectors of the other classes are far
from $\centerTarget$ (prediction value $\indexedPredicted{\output}k$
close to $\sqrt{n}/2=2$). In this spirit, regarding the loss defined
in Equation~\ref{eq:loss-ghost}, we choose $L_{0}^{k}(\indexedPredicted{\output}{\classIndex})$
and $L_{1}^{k}(\indexedPredicted{\output}{\classIndex})$ such that,
for all $\classIndex$:

\begin{equation}
L_{0}^{k}(\indexedPredicted{\output}{\classIndex})=0.5\,\max(0,\indexedPredicted{\output^{2}}{\classIndex}-1.9\,\indexedPredicted{\output}k+0.9)\comma\label{eq:ghostloss0}
\end{equation}
and

{\small{}
	\begin{equation}
	L_{1}^{k}(\indexedPredicted{\output}{\classIndex})=\max(0,\indexedPredicted{\output^{2}}{\classIndex}-0.1\,\indexedPredicted{\output}k)\dot\label{eq:ghostloss1}
	\end{equation}
}{\small\par}

\begin{figure}
\includegraphics[width=0.94\columnwidth]{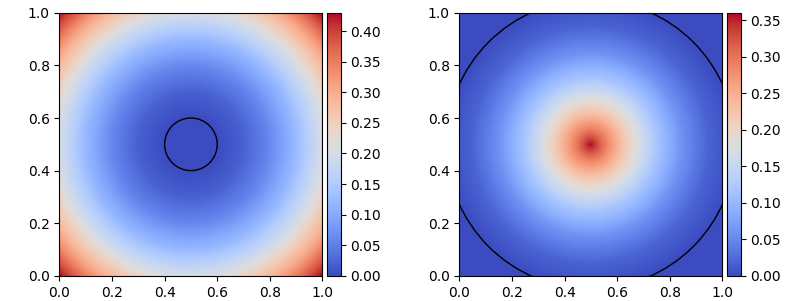}\\
\hspace*{\fill}{\footnotesize{}(a)}\hspace*{\fill}\hspace*{\fill}{\footnotesize{}(b)
	}\hspace*{\fill}\caption{Illustration of the type of loss used in our experiments, in the 2-dimensional
	case ($\protect\dimension=2$). The loss associated with the feature
	vector of the true class is given by plot (a). The loss-free hit zone
	is the area within the black circle. The loss generated by the other
	feature vectors is given by plot (b). The loss-free miss zone is the
	area outside the black circle. \label{fig:mylosses} }
\end{figure}

\noindent The loss associated with the feature vector of the true
class ($\loss_{1}$) and the loss associated with the other feature
vectors ($\loss_{0}$) are represented in Figure~\ref{fig:mylosses}
in the case where $\dimension=2$. The zones where the losses equal
zero are called hit zones (for $L_{1}$) and miss zones (for $L_{0}$)
and are reached when $\indexedPredicted{\output}k<0.1$ and $\indexedPredicted{\output}k>0.9$
respectively. The network has to try to place the feature vector of
the true class in the hit zone of its own space, and the feature vectors
of the other classes in the miss zones of their respective spaces.
For the classification task, the label predicted by $\ourModel$ is
the index of the lowest entry of $\predicted y$.

Besides, a decoder follows the HoM layer to help visualize the results,
as represented in Figure~\ref{fig:Hitnet},
but is not mandatory for the ghost loss to work. During the training,
all the feature vectors of HoM are masked (set to $0$) except the
one related to the true class, then they are concatenated and sent
to the decoder, which produces an output image $\reconstructed{\trainingImage}$,
that aims at reconstructing the initial image $\trainingImage$. The decoder consists in two fully connected
layers of size $512$ and $1024$ with ReLU activations, and one fully
connected layer to a tensor with the same dimensions as the input
image, with a sigmoid activation. The quality of the reconstructed
image $\reconstructed{\trainingImage}$ is evaluated through the mean
squared error with $\trainingImage$. This error is down-weighted
with a multiplicative factor $\alpha$ set to $0.392$ and is then
added to $\loss_{\text{ghost}}$ to produce the final composite loss.

In the following, the networks are run during $250$ epochs with the
Adam optimizer with a constant learning rate of $0.001$, with batches
of $128$ images.

\subsection{Proof of concept}

\begin{figure}
	\begin{centering}
		\includegraphics[width=0.95\columnwidth]{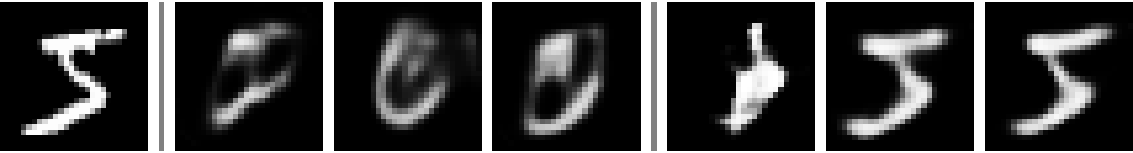}\\
		{\footnotesize{}\hspace*{\fill}~~~(1)\hspace*{\fill}\hspace*{\fill}~~(2)~~\hspace*{\fill}\hspace*{\fill}(3)~~~~~~\hspace*{\fill}(4)~~~~\hspace*{\fill}\hspace*{\fill}(5)~~~~~\hspace*{\fill}(6)~~~~\hspace*{\fill}~~(7)~~\hspace*{\fill}}{\footnotesize\par}
		\par\end{centering}
	\centering{}\caption{(1) Initial image of a digit ``5'', that we mislabel as ``0''.
		Then, the reconstructions from feature vector ``0'' when no ghost
		feature vector is allowed (2), one ghost feature vector is allowed
		for this image only (3), one ghost feature vector is allowed for each
		training image (4). None of the reconstructions based on the feature
		vector of the ``true class 0'' matches the initial image. Then,
		the corresponding reconstructions from feature vector ``5'' (5,6,7).
		When ghost feature vectors are allowed, the network is able to reconstruct
		the initial image from the feature vector ``5'', which corresponds
		to a non-true class, while it is not the case when no ghost feature vector
		is allowed. This indicates that the introduction of ghost feature
		vectors enables the network to learn representative features of non-true
		classes, when necessary.\label{fig:poc-ghost}}
\end{figure}

\begin{figure}
\begin{centering}
\includegraphics[width=0.98\columnwidth]{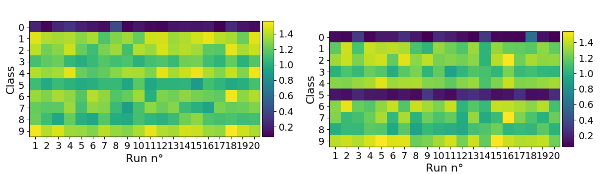}\\
\hspace*{\fill}{\footnotesize{}(a) no ghost feature vector}\hspace*{\fill}\hspace*{\fill}{\footnotesize{}(b)
	one ghost feature vector}\hspace*{\fill}
\par\end{centering}
\caption{Prediction values of the $20$ runs for the erroneously labeled image
when (a) no ghost feature vector is allowed (the network is confident
that the image is a ``0'' and nothing else), and when (b) a ghost
feature vector is allowed for this image. For (b), the network consistently
suspects that the image might actually display the digit ``5''.
\label{fig:poc-preds}}
\end{figure}

In order to illustrate the principle of the ghost loss detailed above,
we perform the following ``proof of concept'' experiment. We consider
the MNIST training set for clearer visual representations. We assign
the label ``0'' to the first image, which actually depicts a digit
``5'', as shown in Figure~\ref{fig:poc-ghost}. Then, we run $\ourModel$
without using the ghost feature vectors nor the ghost loss. At the
end of the training, we compute the prediction values of that image.
In the present setting, let us recall that we expect small prediction
values for the more legitimate classes and large prediction values
for the non plausible classes. As expected, the lowest prediction
value corresponds to the class ``0'', with $\indexedPredicted{\output}0=0.22$;
the image is thus `correctly' classified by the network, which is
an obvious case of overfitting. 
The second lowest prediction value corroborates our hypothesis since
it corresponds to the class ``5'', with $\indexedPredicted{\output}5=1.10$.
Given that $\indexedPredicted{\output}5>0.9$, the network managed
to place the feature vector ``5'' in the miss zone, despite the
fact that the image actually represents a ``5''. The decoder can
be used to visualize the information contained in feature vectors
``0'' and ``5'', as shown in Figure~\ref{fig:poc-ghost}. It
appears that, even though the image is correctly classified as a ``0''
and the network was trained to reconstruct the image from feature
vector ``0'', the reconstruction based on feature vector ``0''
does not match the initial image. Of course, the reconstruction based
on feature vector ``5'' does not represent a proper ``5'' either.
The experiment is repeated several times. In total, $20$ runs of
$\ourModel$ are performed, consistently leading to similar conclusions,
as it can be visualized in Figure~\ref{fig:poc-preds}(a). 

Then, we run the same experiment, but we allow one ghost feature vector
via the ghost loss only for that erroneously labeled image. We do
not impose that this ghost feature vector has to be feature vector
``5''; we simply allow the network, at each epoch, to choose one
feature vector per image that will not generate any loss. At the end
of the training, we compute the prediction values of that image. In
this case, we obtain $\indexedPredicted{\output}0=0.10$ and $\indexedPredicted{\output}5=0.14$,
which indicates that the network does not trust the label provided
since feature vector ``5'', obviously chosen as ghost feature vector,
is close to the hit zone (prediction $<0.1$). All the remaining feature
vectors are in the miss zone (prediction $>0.9$) as expected. In
this case, reconstructing the image from the information contained
in feature vector ``0'' is again far from the initial image, while
the reconstruction from feature vector ``5'' is close to the initial
image (see Figure~\ref{fig:poc-ghost}). Again, the experiment is
run $20$ times. It appears that the ghost feature vector selected
by the network is feature vector ``5'' for the $20$ runs and that
it is consistently close to (and occasionally inside) the hit zone,
with a mean prediction value of $0.17$, as reflected in Figure~\ref{fig:poc-preds}(b).
In fact, the feature vector ``5'' is even closer to the hit zone
than the feature vector ``0'' in $10$ runs. 

Finally, we run the same experiment, but we allow one ghost feature
vector per training image, as if we were in the realistic situation
of not being aware of that erroneous label. The results are similar
to the previous case. The reconstruction from feature vector ``0''
is bad while the one from feature vector ``5'' is good (see Figure~\ref{fig:poc-ghost}).
Among $20$ runs, the ghost feature vector is always feature vector
``5'', with a mean prediction value of $0.14$, and it is closer
to the hit zone than the feature vector ``0'' in $7$ runs. 

\subsection{Automatic detection of abnormalities in well-known training sets}

In the following experiments, we allow one ghost feature vector for
each image of the training set via the ghost loss. This enables us
to analyze the dataset and detect possibly mislabeled images by examining
the ghost feature vectors, their associated ghost classes and ghost
predictions. Let us recall that, for each experience, we trained $\ourModel$
$20$ times, which gives us as many models, and we examine the $20$
ghost feature vectors associated with each image by these models.

\subsubsection{Analysis of MNIST}

First, we study the agreement between the models about the $20$ ghost
classes selected for each image on the MNIST dataset, which is composed
of images representing handwritten digits from ``$0$'' to ``$9$''.
For that purpose, we examine the distribution of the number of different
ghost classes given to each image. It appears that the $20$ models
all agree on the same ghost class for $16.859$ training images, which
represents $28\%$ of the training images. These images have a pair
$\text{(true class, ghost class)}$ and their distribution suggests
that some pairs are more likely to occur, such as $(3,5)$ for $2333$
images, $(4,9)$ for $2580$ images, $(7,2)$ for $1360$ images,
$(9,4)$ for $2380$ images, which gives a glimpse of the classes
that might be likely to be mixed up by the models. Confusions may
occur because of errors in the true labels, but since these numbers
are obtained on the basis of an agreement of $20$ models trained
from the same network structure, this may also indicate the limitations
of the network structure itself in its ability to identify the different
classes. However, a deeper analysis is needed to determine if these
numbers indicate a real confusion or not, that is, which of them make
hits (ghost prediction $<0.1$) and which do not. We can examine if
the $20$ models agree on the ghost predictions of the corresponding
feature vectors. For that purpose, for each image, we compute the
mean and the standard deviation of its $20$ ghost predictions. An
interesting and important observation is that when the mean ghost
prediction gets closer to the hit zone threshold $m$, then the standard
deviation decreases, which indicates that all the models tend to agree
on the fact that a hit is needed.

We can now narrow down the analysis to the images that are the most
likely to be mislabeled, that is, those with a mean ghost prediction
smaller than $0.1$; there are $71$ such images left, which provides
the sanity matrix given in Table~\ref{tab:dist-fs-true-hits}. From
the expected confusions mentioned above, that is $(3,5)$, $(4,9)$,
$(7,2)$, $(9,4)$, we can see that $(7,2)$ is not so much represented
in the present case, while $(1,7)$ and $(7,1)$ subsisted in a larger
proportion, and that $(4,9)$ and $(9,4)$ account for almost half
of the images. A last refinement to our analysis consists of looking
at the number of hits among the $20$ ghost vectors of these $71$
images. It appears that all these images have at least $55\%$ $(11/20)$
of their ghost feature vectors in the hit zone and that more than
$75\%$ $(55/71)$ of the images have a hit for at least $75\%$ $(15/20)$
of the models, which indicates that when the ghost prediction is smaller
than $0.1$, it is the result of a strong agreement between the models.

\begin{table}
\centering{}%
\begin{tabular}{cc|cccccccccc}
&  & \multicolumn{10}{c}{True class}\tabularnewline
&  & 0 & 1 & 2 & 3 & 4 & 5 & 6 & 7 & 8 & 9\tabularnewline
\hline 
\multirow{10}{*}{\rotatebox{90}{Ghost class}} & 0 & 0 & 0 & 0 & 0 & 0 & 0 & 0 & 0 & 0 & 0\tabularnewline
& 1 & 0 & 0 & 0 & 0 & \textbf{\textcolor{blue}{2}} & 0 & 0 & \textbf{\textcolor{blue}{6}} & 0 & 0\tabularnewline
& 2 & 0 & 0 & 0 & 0 & 0 & 0 & 0 & \textbf{\textcolor{blue}{1}} & 0 & 0\tabularnewline
& 3 & 0 & 0 & 0 & 0 & 0 & \textbf{\textcolor{blue}{2}} & 0 & 0 & 0 & 0\tabularnewline
& 4 & 0 & 0 & 0 & 0 & 0 & 0 & \textbf{\textcolor{blue}{1}} & 0 & 0 & \textbf{\textcolor{blue}{20}}\tabularnewline
& 5 & 0 & 0 & 0 & \textbf{\textcolor{blue}{8}} & 0 & 0 & \textbf{\textcolor{blue}{1}} & 0 & 0 & 0\tabularnewline
& 6 & 0 & 0 & 0 & 0 & 0 & \textbf{\textcolor{blue}{2}} & 0 & 0 & 0 & 0\tabularnewline
& 7 & 0 & \textbf{\textcolor{blue}{7}} & \textbf{\textcolor{blue}{2}} & 0 & \textbf{\textcolor{blue}{1}} & 0 & 0 & 0 & 0 & \textbf{\textcolor{blue}{2}}\tabularnewline
& 8 & 0 & 0 & 0 & \textbf{\textcolor{blue}{1}} & 0 & 0 & 0 & 0 & 0 & 0\tabularnewline
& 9 & 0 & 0 & 0 & \textbf{\textcolor{blue}{1}} & \textbf{\textcolor{blue}{13}} & 0 & 0 & \textbf{\textcolor{blue}{1}} & 0 & 0\tabularnewline
\end{tabular}\caption{Sanity matrix for MNIST training images having a unique ghost class
and their mean ghost prediction smaller than the hit zone threshold.}
\label{tab:dist-fs-true-hits}
\end{table}

\begin{figure}
	\centering{}\includegraphics[width=0.9\columnwidth]{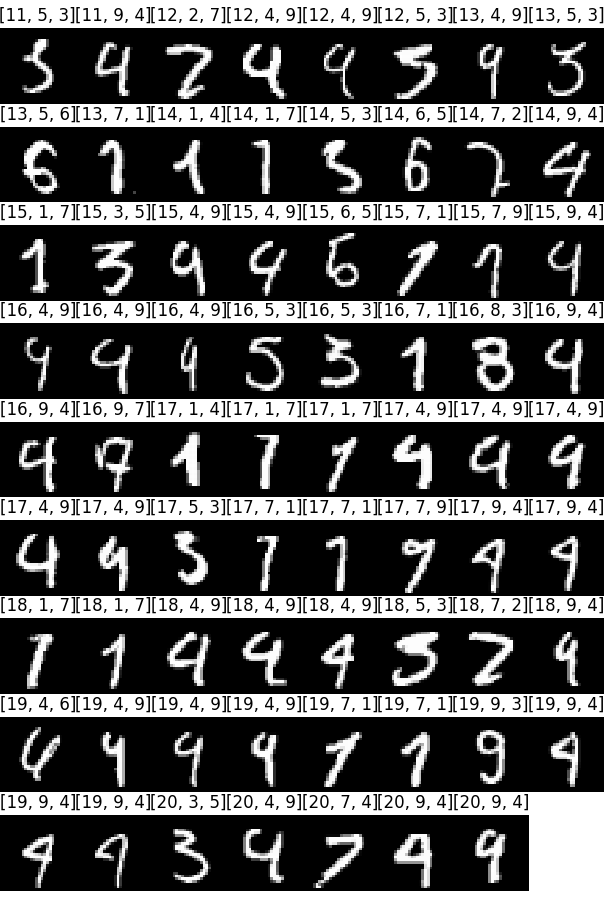}\caption{The $71$ images of MNIST training set whose $20$ ghost feature vectors
		have a mean ghost prediction smaller than the hit zone threshold,
		sorted by increasing number of hits. A triplet $[\mathtt{a},\mathtt{b},\mathtt{c}]$
		indicates that $\mathtt{a}$ hits have been observed in the ghost
		class $\mathtt{b}$ whereas the true class is $\mathtt{c}$.\label{fig:all-monsters}}
\end{figure}

Finally, the $71$ images, sorted by number of hits, are represented
in Figure~\ref{fig:all-monsters}. The number of hits, the ghost
class, and the true class are indicated for each image. Some of these
images are clearly mislabeled, such as the image with the first $[16,5,3]$
tag and the one with the $[20,7,4]$ tag. Others are terribly confusing
by looking almost the same but having different true classes, such
as the images with the $[12,2,7]$ and $[18,7,2]$ tags, which explains
the ghost classes selected. While pursuing a different purpose, the
DropMax technique used in~\cite{Lee2017DropMax} allowed the authors
to identify ``hard cases'' of the training set which are among the
$71$ images represented in Figure~\ref{fig:all-monsters}. Some
misleading images in the test set, presenting similar errors as those
shown in Figure~\ref{fig:all-monsters}, are displayed in \eg~\cite{Mayraz2002Recognizing}.
All in all, MNIST training set appears to be globally reliable, even
though a few images could be either removed or relabeled. 

\subsubsection{Analysis of Fashion-MNIST}

\begin{figure}
	\begin{centering}
		\includegraphics[width=0.95\columnwidth]{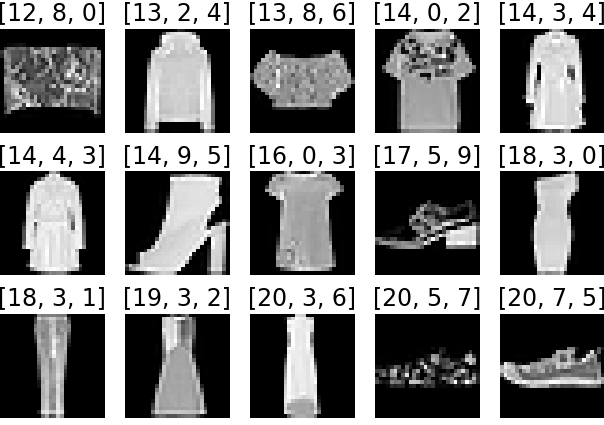}\caption{A selection of training images of Fashion-MNIST whose $20$ ghost
			feature vectors have a mean ghost prediction smaller than the hit
			zone threshold.  The classes are: T-shirt/top (0), trouser (1), pullover
			(2), dress (3), coat (4), sandal (5), shirt (6), sneaker (7), bag
			(8), ankle boot (9).}
		\label{fig:monsters-fashion}
		\par\end{centering}
\end{figure}

Zalando's Fashion-MNIST dataset is a MNIST-like dataset in the fashion
domain. The images are divided in ten classes: T-shirt/top (0), trouser
(1), pullover (2), dress (3), coat (4), sandal (5), shirt (6), sneaker
(7), bag (8), ankle boot (9). We proceed exactly as above to analyze
this dataset with ghost feature vectors. It appears that the $20$
models all agree on the same ghost class for $39074$ training images,
which represents $65\%$ of the training images. This large percentage
can be seen as a first warning sign of frequent confusions between
classes. Among them, $7848$ have a mean ghost prediction smaller
than $0.1$, which still represents more than $13\%$ of the training
set. The sanity matrix can be found in Table~\ref{tab:dist-ghost-fashion}
and a few of these images are displayed in Figure~\ref{fig:monsters-fashion}.
Let us also note that $1123$ of them display $20$ hits in their
ghost class. These results indicate that the classes of Fashion-MNIST
may not be as well separated as in the case of MNIST. For example,
from Table~\ref{tab:dist-ghost-fashion}, it can be inferred that
the classes ``T-shirt/top'' and ``shirt'' are likely to be mixed
up by the network, the same observation holds for ``sandal'' and
``sneaker'' or ``ankle boot'' and ``sneaker'' quite rightly.
Such proposed alternatives are not particularly surprising when looking
at the images displayed in Figure~\ref{fig:monsters-fashion}, which
are hardly distinguishable with the naked eye. Interestingly, many
``trouser'' images have their ghost feature vectors in the ``dress''
class, which comes from the fact that many images of dresses are simply
represented by a white vertical pattern, as for some trousers. These
observations raise the question of the quality of the dataset. While
it is possible that some confusions can be handled with larger and
deeper networks, a visual inspection of the images identified as misleading
in this first study reveals that this classification problem might
be partially ill-posed, in the sense that several categories could
be merged or two labels could be assigned to many images. Alternatively,
one can argue that the resolution of the images is simply too low
to properly identify the correct class, and that a much more interesting
problem would be the classification of higher resolution images of
fashion items. In any case, artificial difficulties arise with Fashion-MNIST
dataset, which are probably present in the test set as well. Fashion-MNIST
may thus not necessarily be used as a ``direct drop-in replacement
for the original MNIST dataset for benchmarking machine learning algorithms''
as openly intended; it may rather serve as a valuable complementary
dataset. 

\begin{table}
\centering{}{\scriptsize{}}%
\begin{tabular*}{0.99\columnwidth}{@{\extracolsep{\fill}}>{\centering}p{0.1cm}>{\centering}p{0.1cm}|>{\centering}m{0.1cm}>{\centering}m{0.1cm}>{\centering}m{0.1cm}>{\centering}m{0.1cm}>{\centering}m{0.1cm}>{\centering}m{0.1cm}>{\centering}m{0.1cm}>{\centering}m{0.1cm}>{\centering}m{0.1cm}>{\centering}m{0.1cm}}
&  & \multicolumn{10}{c}{{\scriptsize{}True class}}\tabularnewline
&  & {\scriptsize{}0} & {\scriptsize{}1} & {\scriptsize{}2} & {\scriptsize{}3} & {\scriptsize{}4} & {\scriptsize{}5} & {\scriptsize{}6} & {\scriptsize{}7} & {\scriptsize{}8} & {\scriptsize{}9}\tabularnewline
\hline 
\multirow{10}{0.1cm}{{\scriptsize{}\rotatebox{90}{Ghost class}}} & {\scriptsize{}0} & {\scriptsize{}0} & {\scriptsize{}0} & \textcolor{blue}{\scriptsize{}4} & \textcolor{blue}{\scriptsize{}45} & {\scriptsize{}0} & {\scriptsize{}0} & \textcolor{blue}{\scriptsize{}994} & {\scriptsize{}0} & {\scriptsize{}0} & {\scriptsize{}0}\tabularnewline
& {\scriptsize{}1} & {\scriptsize{}0} & {\scriptsize{}0} & {\scriptsize{}0} & \textcolor{blue}{\scriptsize{}83} & {\scriptsize{}0} & {\scriptsize{}0} & {\scriptsize{}0} & {\scriptsize{}0} & {\scriptsize{}0} & {\scriptsize{}0}\tabularnewline
& {\scriptsize{}2} & {\scriptsize{}0} & {\scriptsize{}0} & {\scriptsize{}0} & {\scriptsize{}0} & \textcolor{blue}{\scriptsize{}138} & {\scriptsize{}0} & \textcolor{blue}{\scriptsize{}46} & {\scriptsize{}0} & {\scriptsize{}0} & {\scriptsize{}0}\tabularnewline
& {\scriptsize{}3} & \textcolor{blue}{\scriptsize{}35} & \textcolor{blue}{\scriptsize{}1004} & \textcolor{blue}{\scriptsize{}4} & {\scriptsize{}0} & \textcolor{blue}{\scriptsize{}156} & {\scriptsize{}0} & \textcolor{blue}{\scriptsize{}9} & {\scriptsize{}0} & {\scriptsize{}0} & {\scriptsize{}0}\tabularnewline
& {\scriptsize{}4} & {\scriptsize{}0} & {\scriptsize{}0} & \textcolor{blue}{\scriptsize{}185} & \textcolor{blue}{\scriptsize{}33} & {\scriptsize{}0} & {\scriptsize{}0} & \textcolor{blue}{\scriptsize{}95} & {\scriptsize{}0} & \textcolor{blue}{\scriptsize{}1} & {\scriptsize{}0}\tabularnewline
& {\scriptsize{}5} & {\scriptsize{}0} & {\scriptsize{}0} & {\scriptsize{}0} & {\scriptsize{}0} & {\scriptsize{}0} & {\scriptsize{}0} & {\scriptsize{}0} & \textcolor{blue}{\scriptsize{}1220} & {\scriptsize{}0} & \textcolor{blue}{\scriptsize{}13}\tabularnewline
& {\scriptsize{}6} & \textcolor{blue}{\scriptsize{}1131} & {\scriptsize{}0} & \textcolor{blue}{\scriptsize{}60} & {\scriptsize{}0} & \textcolor{blue}{\scriptsize{}51} & {\scriptsize{}0} & {\scriptsize{}0} & {\scriptsize{}0} & {\scriptsize{}0} & {\scriptsize{}0}\tabularnewline
& {\scriptsize{}7} & {\scriptsize{}0} & {\scriptsize{}0} & {\scriptsize{}0} & {\scriptsize{}0} & {\scriptsize{}0} & \textcolor{blue}{\scriptsize{}1105} & {\scriptsize{}0} & {\scriptsize{}0} & {\scriptsize{}0} & \textcolor{blue}{\scriptsize{}801}\tabularnewline
& {\scriptsize{}8} & \textcolor{blue}{\scriptsize{}2} & {\scriptsize{}0} & {\scriptsize{}0} & {\scriptsize{}0} & {\scriptsize{}0} & {\scriptsize{}0} & \textcolor{blue}{\scriptsize{}2} & {\scriptsize{}0} & {\scriptsize{}0} & {\scriptsize{}0}\tabularnewline
& {\scriptsize{}9} & {\scriptsize{}0} & {\scriptsize{}0} & {\scriptsize{}0} & {\scriptsize{}0} & {\scriptsize{}0} & \textcolor{blue}{\scriptsize{}34} & {\scriptsize{}0} & \textcolor{blue}{\scriptsize{}597} & {\scriptsize{}0} & {\scriptsize{}0}\tabularnewline
\end{tabular*}\caption{Sanity matrix for Fashion-MNIST training images having a unique ghost
class and their mean ghost prediction smaller than the hit zone threshold.
The classes are: T-shirt/top (0), trouser (1), pullover (2), dress
(3), coat (4), sandal (5), shirt (6), sneaker (7), bag (8), ankle
boot (9).}
\label{tab:dist-ghost-fashion}
\end{table}

\subsubsection{Analysis of SVHN}

\begin{figure}
	
	\centering{}\includegraphics[width=0.95\columnwidth]{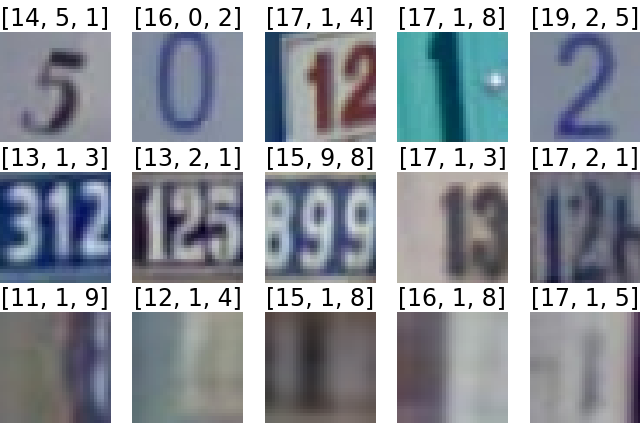}\caption{A selection of training images of SVHN whose $20$ ghost feature vectors
		have a mean ghost prediction smaller than the hit zone threshold.
		The three rows represent images with different types of peculiarities.
		A triplet $[\mathtt{a},\mathtt{b},\mathtt{c}]$ indicates that $\mathtt{a}$
		hits have been observed in the ghost class $\mathtt{b}$ whereas the
		true class is $\mathtt{c}$.}
	\label{fig:monsters-svhn}
\end{figure}

Following the same process as above, we use ghost feature vectors
to analyze the SVHN dataset. It appears that the $20$ models all
agree on the same ghost class for $15516$ training images, which
represents $21\%$ of the training images. Among them, $121$ have
a mean ghost prediction smaller than $0.1$. A few of them are displayed
in Figure~\ref{fig:monsters-svhn}. Several types of abnormalities
can be detected with ghost feature vectors in this case. First, we
are able to detect training images that are clearly mislabeled as
in MNIST, regardless of the number of digits actually displayed in
the image, as shown in the first row of Figure~\ref{fig:monsters-svhn}.
These errors may originate from the automatic annotation process~\cite{Netzer2011Reading}.
For example, the images with the tags $[16,0,2]$ and $[19,2,5]$
are actually crops of a same larger image, which presumably had the
correct full house number encoded but for which the cropping procedure
may have accidentally missed a digit, resulting in a shift of the
labels of individual digits. Then, the use of ghost feature vectors
allows to detect images that may deserve multiple labels since multiple
digits are present in the images, as seen in the second row of Figure~\ref{fig:all-monsters}.
It is important to note that the true label of most of the training
images displaying at least two digits corresponds to the most centered
digit. Hence, when several digits are present and that the label does
not correspond to the most centered one, it is not surprising to observe
that $\ourModel$ suggests ghost classes corresponding to these centered
digits, as in Figure~\ref{fig:monsters-svhn}. Therefore, one could
also consider these images as cases of mislabeled images rather than
images deserving multiple labels if only one label is allowed. Again,
the automatic procedure used to crop and label the images can be responsible
of these anomalies. Finally, ghost feature vectors allow to pinpoint
some training images that do not seem to actually represent any digit,
as shown in the third row of Figure~\ref{fig:monsters-svhn}. Such
images may rather correspond to some kind of background, possibly
cropped close to actual digits that have been missed in the automatic
annotation process. In the cases displayed in Figure~\ref{fig:monsters-svhn},
we hypothesize that the $20$ models all agree on the ghost class
``1'' because of the kind of change of surface that generates a
vertical pattern in the middle of the images. Several similar images
can also be found among the images whose $20$ ghost feature vectors
are distributed in more than one ghost class. It is important to note
that there are also confusing images among those whose mean ghost
prediction is slightly larger than $0.1$, which extends the number
of debatable training images. Given the fact that the test set commonly
used for assessing the performances of models comes from the same
database, it is likely to contain similar misleading images as well,
which somehow sets an upper bound on the best performance achievable
on this dataset. Consequently, if one only looks at test error rates
reported by the models, this dataset might actually present an artificial
apparent difficulty. The test set should thus be carefully examined
and maybe corrected to have a better idea of the true performance
of the models.

\subsubsection{Analysis of CIFAR10}

Among the training sets analyzed with the use of ghost feature vectors,
CIFAR10 seems to be the most reliable. Indeed, using $\ourModel$,
it appears that the $20$ models all agree on the same ghost class
for $10728$ training images, which represents $21\%$ of the training
images. However, among them, only $15$ have a mean ghost prediction
smaller than $0.1$, and a manual check showed that these images are
correctly labeled. Most of them $(11)$ are birds or boats whose ghost
feature vectors correspond to the airplane class, which is presumably
due to their common blue background. We also extended the threshold
and checked dozens of images with a mean ghost prediction smaller
than $0.2$ (there are $1812$ of them) but we did not find any clearly
mislabeled images. Among the images with several ghost classes, those
with the most significant predictions did not appear controversial
either. Our tests, with the limitations of the architecture of $\ourModel$,
tend to indicate that the dataset is correctly annotated. For the
record, using ghost feature vectors in the same way but with a DenseNet-40-12~\cite{Huang2017Densely}
backbone network before the HoM layer does not reveal misleading images
either. It even appears that the $20$ models do not agree on a single
ghost class for any image, which tends to confirm that CIFAR10 is
a reliable dataset.

\section{Conclusion}

We introduce the notion of ``ghost'', which allows the network
to select an alternative choice to the label provided in a classification
task without being penalized during the training phase and thus helps
detecting abnormalities in training images. This is done by zeroing out the loss associated with the most likely incorrect prediction, which enables the network to build features in line with the correct label of mislabeled samples.
After illustrating the
use of the ghost loss in a proof of concept experiment, we use it
as a tool for analyzing well-known training sets (MNIST, Fashion-MNIST,
SVHN, CIFAR10) and detect possible anomalies therein via the novel
concept of ``sanity matrix''. As far as MNIST is concerned, a few
training images that are clearly mislabeled are detected, as well
as pairs of images that are almost identical but have different labels,
which indicates that the dataset is globally reliable. Regarding Fashion-MNIST,
it appears that a large number of training data is considered misleading
in some way by the network, suggesting that some classes could be
merged, or that some images could deserve multiple labels or that
higher resolution images could be provided to define a better classification
problem. In SVHN, we find that several dozens of images are either
mislabeled, deserve multiple labels, or do not represent any digit,
presumably because of the automatic cropping-based labelling process.
The CIFAR10 dataset appears to be the most reliable since no particular
anomaly is detected. The ghost loss, in combination with a dedicated
neural network, could thus be used in the design of a reliability
measure for the images of a training set.

\paragraph*{Acknowledgements.}

This work was supported by the DeepSport project of the Walloon region (Belgium). A. Cioppa has a grant funded by the FRIA (Belgium).


\begin{thebibliography}{00}

\bibitem{Brodley1999Identifying}
C.~Brodley and M.~Friedl.
\newblock Identifying mislabeled training data.
\newblock {\em Journal of Artificial Intelligence Research}, 11:131--167,
August 1999.

\bibitem{Deliege2018HitNet}
A.~Deli\`ege, A.~Cioppa, M.~Van~Droogenbroeck.
\newblock {HitNet}: a neural network with capsules embedded in a Hit-or-Miss layer, extended with hybrid data augmentation and ghost capsules.
\newblock {\em CoRR}, abs/1806.06519, 2018.

\bibitem{Deng2009ImageNet}
J.~Deng, W.~Dong, R.~Socher, L.~Li, K.~Li, and L.~Fei-Fei.
\newblock {ImageNet}: A large-scale hierarchical image database.
\newblock In {\em IEEE International Conference on Computer Vision and Pattern
	Recognition (CVPR)}, pages 248--255, Miami, Florida, USA, June 2009.

\bibitem{Frenay2014Classification}
B.~Frenay and M.~Verleysen.
\newblock Classification in the presence of label noise: A survey.
\newblock {\em IEEE Transactions on Neural Networks and Learning Systems},
25(5):845--869, May 2014.

\bibitem{quickdraw}
{Google Creative Lab}.
\newblock Quick, draw!
\newblock \url{https://quickdraw.withgoogle.com/data}, 2017.

\bibitem{Huang2017Densely}
G.~Huang, Z.~Liu, L.~{van der Maaten}, and K.~Weinberger.
\newblock Densely connected convolutional networks.
\newblock In {\em IEEE International Conference on Computer Vision and Pattern
	Recognition (CVPR)}, pages 2261--2269, Honolulu, Hawaii, USA, July 2017.

\bibitem{Karimi2020DeepLearning}
D.~Karimi, H.~Dou, S.~K.~Warfield, A.~Gholipour.
\newblock Deep learning with noisy labels: exploring techniques and remedies in medical image analysis
\newblock {\em CoRR}, abs/1912.02911 (v2), 2020.

\bibitem{Krizhevsky2009Learning}
A.~Krizhevsky.
\newblock Learning multiple layers of features from tiny images, 2009.
\newblock Technical report, University of Toronto.

\bibitem{LeCun2001Gradient}
Y.~Lecun, L.~Bottou, Y.~Bengio, and P.~Haffner.
\newblock Gradient-based learning applied to document recognition.
\newblock {\em Proceedings of IEEE}, 86(11):2278--2324, November 1998.

\bibitem{Lee2017DropMax}
H.~Lee, J.~Lee, S.~Kim, E.~Yang, and S.~Hwang.
\newblock Drop{M}ax: Adaptive variational softmax.
\newblock {\em CoRR}, abs/1712.07834, December 2017.

\bibitem{Mayraz2002Recognizing}
G.~Mayraz and G.~E. Hinton.
\newblock Recognizing handwritten digits using hierarchical products of
experts.
\newblock {\em IEEE Transactions on Pattern Analysis and Machine Intelligence},
24(2):189--197, February 2002.

\bibitem{Netzer2011Reading}
Y.~Netzer, T.~Wang, A.~Coates, A.~Bissacco, B.~Wu, and A.~Ng.
\newblock Reading digits in natural images with unsupervised feature learning.
\newblock In {\em Advances in Neural Information Processing Systems (NIPS)},
volume 2011, Granada, Spain, December 2011.

\bibitem{Patrini2017Making}
G.~Patrini, A.~Rozza, A.~Menon, R.~Nock, and L.~Qu.
\newblock Making deep neural networks robust to label noise: A loss correction
approach.
\newblock In {\em IEEE International Conference on Computer Vision and Pattern
	Recognition (CVPR)}, pages 2233--2241, Honolulu, Hawaii, USA, July 2017.

\bibitem{Sabour2017Dynamic}
S.~Sabour, N.~Frosst, and G.~Hinton.
\newblock Dynamic routing between capsules.
\newblock {\em CoRR}, abs/1710.09829, October 2017.

\bibitem{Srivastava2014Dropout}
N.~Srivastava, G.~Hinton, A.~Krizhevsky, I.~Sutskever, and R.~Salakhutdinov.
\newblock Dropout: A simple way to prevent neural networks from overfitting.
\newblock {\em Journal of Machine Learning Research}, 15(1):1929--1958, January
2014.

\bibitem{Xiao2017Fashion}
H.~Xiao, K.~Rasul, and R.~Vollgraf.
\newblock Fashion-{MNIST}: a novel image dataset for benchmarking machine
learning algorithms.
\newblock {\em CoRR}, abs/1708.07747, 2017.

\bibitem{Zhang2018Generalized}
Z.~Zhang, M.~Sabuncu.
\newblock Generalized cross entropy loss for training deep neural  networks with noisy labels.
\newblock In {\em Advances in Neural Information Processing Systems (NIPS)}, pages 8778-8788, 2018.

\bibitem{Thulasidasan2019Combating}
S.~Thulasidasan, T.~Bhattacharya, J.~Bilmes, G.~Chennupati, J.~Mohd-Yusof.
\newblock Combating Label Noise in Deep Learning Using Abstention.
\newblock In {\em International Conference on Machine Learning (ICML)}, 2019.

\bibitem{Rusiecki2019Trimmed}
A.~Rusiecki.
\newblock Trimmed robust loss function for training deep neural networks  with label noise.
\newblock In {\em International Conference on Artificial Intelligence and Soft Computing}, 2019.

\bibitem{Speth2019Automated}
J.~Speth, E.~M.~Hand.
\newblock Automated label noise identification for facial attribute recognition.
\newblock In {\em IEEE International Conference on Computer Vision and Pattern Recognition Workshops (CVPRW)}, 2019.

\bibitem{Wang2019Emphasis}
X.~Wang, Y.~Hua, E.~Kodirov, and N.= M.~Robertson.
\newblock Emphasis Regularisation by Gradient Rescaling for Training Deep Neural Networks with Noisy Labels.
\newblock {\em CoRR}, abs/1905.11233, 2019.

\bibitem{Lee2018CleanNet}
K.-H.~Lee, X.~He, L.~Zhang, and L.~Yang.
\newblock CleanNet: Transfer Learning for Scalable Image Classifier Training with Label Noise.
\newblock In {\em IEEE International Conference on Computer Vision and Pattern Recognition (CVPR)}, pages 5447–5456, Salt Lake City, Utah, USA, June 2018.



\end{thebibliography}
\end{document}